\documentclass{article}

\usepackage{microtype}
\usepackage{graphicx}
\usepackage{subfigure}
\usepackage{booktabs} % for professional tables
\usepackage{chapterbib}
\usepackage{hyperref}
\usepackage{algpseudocode}

\newcommand{\myparagraph}[1]{
\vspace{0.1cm}\noindent
\textbf{#1.}
}
 \usepackage [accepted]{icml2025}

\usepackage{amsmath}
\usepackage{amssymb}
\usepackage{mathtools}
\usepackage{amsthm}

\usepackage[capitalize,noabbrev]{cleveref}

\theoremstyle{plain}

\theoremstyle{definition}

\theoremstyle{remark}

\usepackage[textsize=tiny]{todonotes}

\icmltitlerunning{Exploring Position Encoding Mechanism in Diffusion U-Net for Training-free High-resolution Image Generation}

\begin{document}

\twocolumn[
\icmltitle{Exploring Position Encoding in Diffusion U-Net for Training-free High-resolution Image Generation}

% It is OKAY to include author information, even for blind
% submissions: the style file will automatically remove it for you
% unless you've provided the [accepted] option to the icml2025
% package.

% List of affiliations: The first argument should be a (short)
% identifier you will use later to specify author affiliations
% Academic affiliations should list Department, University, City, Region, Country
% Industry affiliations should list Company, City, Region, Country

% You can specify symbols, otherwise they are numbered in order.
% Ideally, you should not use this facility. Affiliations will be numbered
% in order of appearance and this is the preferred way.
\icmlsetsymbol{equal}{*}
\icmlsetsymbol{cor}{$\dagger$}
\begin{icmlauthorlist}
\icmlauthor{Feng Zhou}{equal,bupt}
\icmlauthor{Pu Cao}{equal,bupt}
\icmlauthor{Yiyang Ma}{bupt}
\icmlauthor{Lu Yang}{bupt}
\icmlauthor{Jianqin Yin}{cor,bupt}
\end{icmlauthorlist}

\icmlaffiliation{bupt}{Beijing University of Posts and Telecommunications}

\icmlcorrespondingauthor{Feng Zhou}{zhoufeng@bupt.edu.cn}
\icmlcorrespondingauthor{Pu Cao}{caopu@bupt.edu.cn}
\icmlcorrespondingauthor{Jianqin Yin}{jqyin@bupt.edu.cn}
% You may provide any keywords that you
% find helpful for describing your paper; these are used to populate
% the "keywords" metadata in the PDF but will not be shown in the document
\icmlkeywords{Machine Learning, ICML}

\vskip 0.3in
]

% this must go after the closing bracket ] following \twocolumn[ ...

% This command actually creates the footnote in the first column
% listing the affiliations and the copyright notice.
% The command takes one argument, which is text to display at the start of the footnote.
% The \icmlEqualContribution command is standard text for equal contribution.
% Remove it (just {}) if you do not need this facility.

\printAffiliationsAndNotice{\icmlEqualContribution, \icmlCorrespondingAuthor}   % leave blank if no need to mention equal contribution
%\printAffiliationsAndNotice{\icmlEqualContribution} % otherwise use the standard text.

\begin{abstract}
Denoising higher-resolution latents via a pre-trained U-Net leads to repetitive and disordered image patterns. Although recent studies make efforts to improve generative quality by aligning denoising process across original and higher resolutions, the root cause of suboptimal generation is still lacking exploration. Through comprehensive analysis of position encoding in U-Net, we attribute it to inconsistent position encoding, sourced by the inadequate propagation of position information from zero-padding to latent features in convolution layers as resolution increases. To address this issue, we propose a novel training-free approach, introducing a Progressive Boundary Complement (PBC) method. This method creates dynamic virtual image boundaries inside the feature map to enhance position information propagation, enabling high-quality and rich-content high-resolution image synthesis. Extensive experiments demonstrate the superiority of our method. 

\end{abstract}

\section{Introduction}
\begin{figure}[t]
\begin{center}
\centerline{\includegraphics[width=\columnwidth]{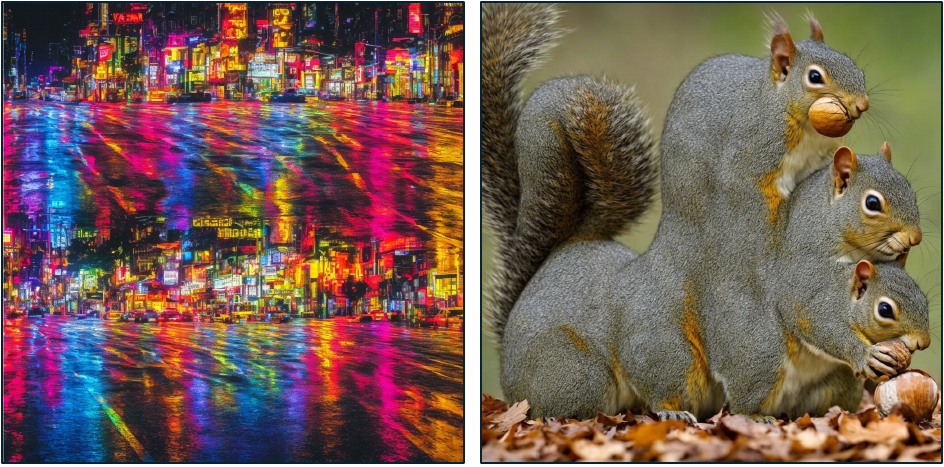}}
\caption{Directly generating high-resolution images using a pre-trained Latent Diffusion Model result in repetitive (\emph{right}) and disordered (\emph{left}) patterns.}
\label{fig-intro-1}
\end{center}
\vskip -0.5in
\end{figure}

Text-to-image generation has gained significant attention due to its wide range of applications in real-world scenarios, such as digital content creation~\cite{poole2022dreamfusion} and personalized media generation~\cite{cao2024controllable}. Among various generative paradigms, Latent Diffusion Models (LDMs)~\cite{rombach2022high} have emerged as a popular and powerful approach, showing impressive results. Traditional LDMs, like Stable Diffusion (SD)~\cite{rombach2022high}, employ U-Net for latent-space denoising with a relatively fixed training resolution (\emph{e.g.}, 64×64 in \emph{SD-2.1}). Recent research~\cite{he2023scalecrafter, jin2023training} shows that the pre-trained U-Net is highly sensitive to the latent resolutions; higher-resolution latent inputs often result in repetitive patterns and disordered layouts, as shown in Figure~\ref{fig-intro-1}. This phenomenon can be described as ``image elements appearing in wrong positions".  In this study, we characterize this disorder as the absence of consistent position encoding during the high-resolution generation in latent features and investigate it from the U-Net's architectures. 

We first look into how U-Net constructs the position information across elements or tokens. Since no positional embeddings are added in the attention layers, all positional information is expected to originate solely from the zero-padding mechanism in the convolution layers ~\cite{islam2020much}. Specifically, 1) the zero-padding at the boundaries of the feature map provides position information, and 2) this position information propagates by multiple stacked convolutional layers in UNet from the boundary into the central part, ultimately influencing the spatial arrangement in the generation. As the feature map size increases, the propagation paths inevitably become longer, resulting in inaccurate position encoding in latent features. This issue essentially degrades generation performance at higher resolution. We further designed both quantitative and comparative experiments to verify the absence of positional information in high-resolution generation and how this issue ultimately affects generation quality. 

Existing methods, based on position information absence perspective, can be regarded as broadly designed to address this challenge in high-resolution generation by align position information propagation across different resolutions. For instance, dilated convolution-based method~\cite{he2023scalecrafter, huang2025fouriscale} leverages dilating the convolution kernel to accelerate the propagation of position information in high-resolution generation. Multi-stage lifting methods~\cite{cao2024ap, qiu2024freescale, kim2024diffusehigh}, which progressively increase image resolution by multiple stages, maintain position information by lifting resolution from the original level. However, these approaches have notable limitations: 1) they adopt an engineering-driven strategy by enforcing positional alignment at higher resolutions, yet lack theoretical justification and often introduce complex architectures. 2) this rigid alignment constrains content diversity across different resolutions, making it difficult to generate more enriched and complex visual details at high resolutions.

In this paper, we propose a simple but effective training-free high-resolution image generation method to tackle this challenge, simplifying the task by directly addressing it from the perspective of position information completion. We introduce a Progressive Boundary Complement (PBC) method, designed to enhance the propagation of position information from the boundary to the central regions. To achieve this, we construct virtual boundaries within the feature map, a specialized form of unidirectional padding, to facilitate position information transmission. By hierarchically placing these virtual boundaries within the feature map, our method not only corrects position encoding inconsistencies in high-resolution features, thereby mitigating spatial distortions but also effectively extends the perceived image boundaries. This enables the generation of high-resolution images with richer details and greater content diversity. Our extensive experiments demonstrate its effectiveness. 

To sum up, our contributions are as follows:
\begin{itemize}
\item We throughly investigate the mechanism of position encoding in diffusion U-Net and reach: the position information originates from zero-padding and propagated to the entire feature map through stacked convolution layers.  Based on this, we identify the root cause of the disordered image patterns in high-resolution image to the inconsistent position encoding, sourced by inadequate propagation of position information as feature map increase. 
\item  To address this issue, we propose Progressive Boundary Complement (PBC), a training-free method that enhances position information propagation by introducing hierarchical virtual boundaries within the feature map.	 It extends the perceived image boundaries, enabling the synthesis of high-resolution images with richer content.
\item Extensive quantitative and qualitative experiments demonstrate the superiority of our method in generating high-resolution images with enhanced content richness.
\end{itemize}
\section{Related Works}
% analysis of early-stage denoising
\subsection{High-Resolution Image Synthesis with Pre-trained Diffusion Models}
Diffusion models~\cite{rombach2022high, peebles2023scalable} have significantly propelled the development of text-to-image generation. However, as these models are trained at relatively fixed resolutions, they struggle to effectively generate higher-resolution images. Several methods address this limitation by fine-tuning pre-trained models, converting them into dedicated high-resolution image generators~\cite{du2024demofusion, guo2025make}. In contrast, many approaches tackle this issue without additional training~\cite{hwang2024upsample, huang2025fouriscale}. For instance, certain studies have explored adapting model architectures specifically for high-resolution generation. Attn-Entro~\cite{jin2023training} adjusts the attention scaling factor based on attention entropy considerations during high-resolution generation. ScaleCrafter~\cite{he2023scalecrafter} increases the receptive field of convolutional layers to facilitate high-resolution image synthesis. HiDiffusion~\cite{zhang2025hidiffusion} dynamically modifies feature map dimensions to mitigate object duplication.
Another category of training-free methods adopts a multi-stage resolution-lifting paradigm, initially generating images at the base resolution and progressively upsampling them to the desired higher resolutions~\cite{cao2024ap, kim2024diffusehigh, wu2024megafusion, yang2024fam, zhang2024frecas, qiu2024freescale}. Additionally, some methods emphasize refining text prompts at the patch level to enhance high-resolution image generation~\cite{bar2023multidiffusion, lin2025accdiffusion, liu2024hiprompt}.

\subsection{Mechanisms of Text-to-Image Diffusion Models.}
Although text-to-image diffusion models have achieved remarkable success across various applications, their internal mechanisms have not yet been fully explored~\cite{si2024freeu}. For instance, FreeU~\cite{si2024freeu} analyzes the roles played by the U-Net’s backbone and skip connections from a frequency-domain perspective. Additionally, several studies have examined properties of text encoders and cross-attention mechanisms, shedding light on the interactions between textual prompts and the diffusion process~\cite{toker2024diffusion, yang2024diffusion, yi2024towards}. In this study, we aim to further investigate the underlying mechanisms from the perspective of positional information construction of U-Net.
\begin{figure}[t]
\begin{center}
\centerline{\includegraphics[width=\columnwidth]{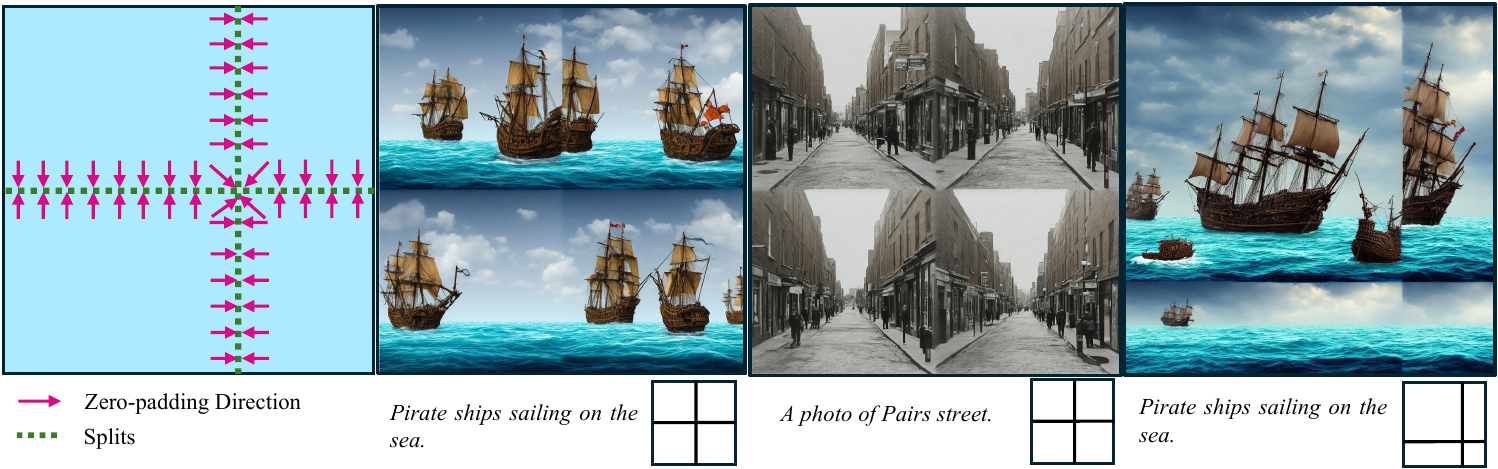}}
\vskip -0.1in
\caption{Trench-style Zero-padding Technique. The left-side graph diagram illustrates the process of applying bi-directional zero-padding to the feature map in the convolution operation. The three images on the right show 1024×1024 resolution outputs using this technique, with the corresponding split sketches displayed below.}
\label{fig-analysis-1}
\end{center}
\vskip -0.3in
\end{figure}

\begin{figure}[t]
\begin{center}
\centerline{\includegraphics[width=\columnwidth]{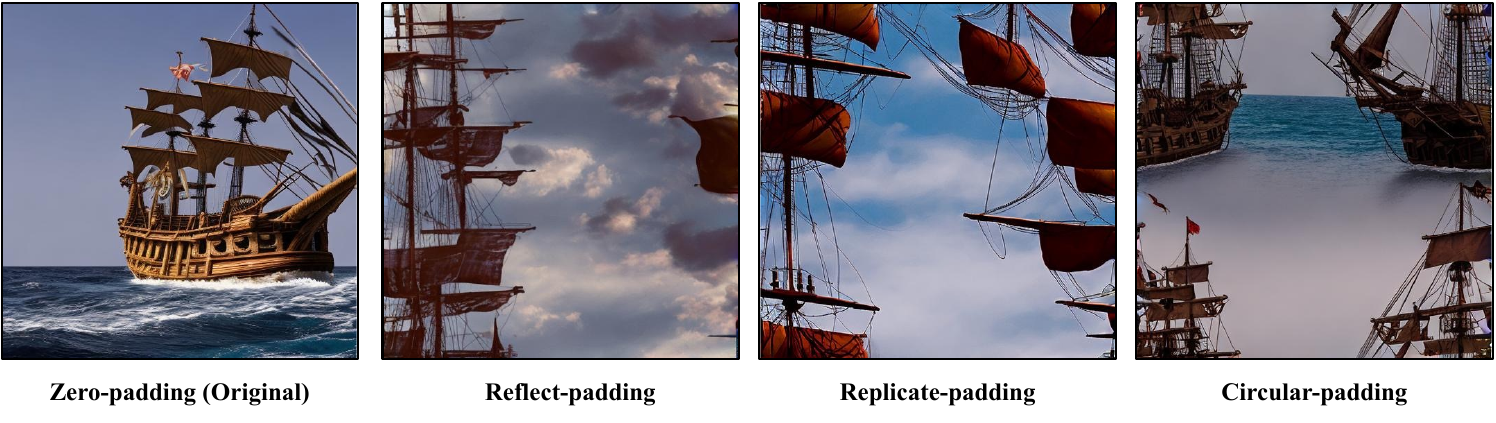}}
\vskip -0.1in
\caption{Padding Type Analysis. We evaluate the effect of different padding types.}
\label{fig-analysis-2}
\end{center}
\vskip -0.3in
\end{figure}

\begin{figure}[t]
\begin{center}
\centerline{\includegraphics[width=\columnwidth]{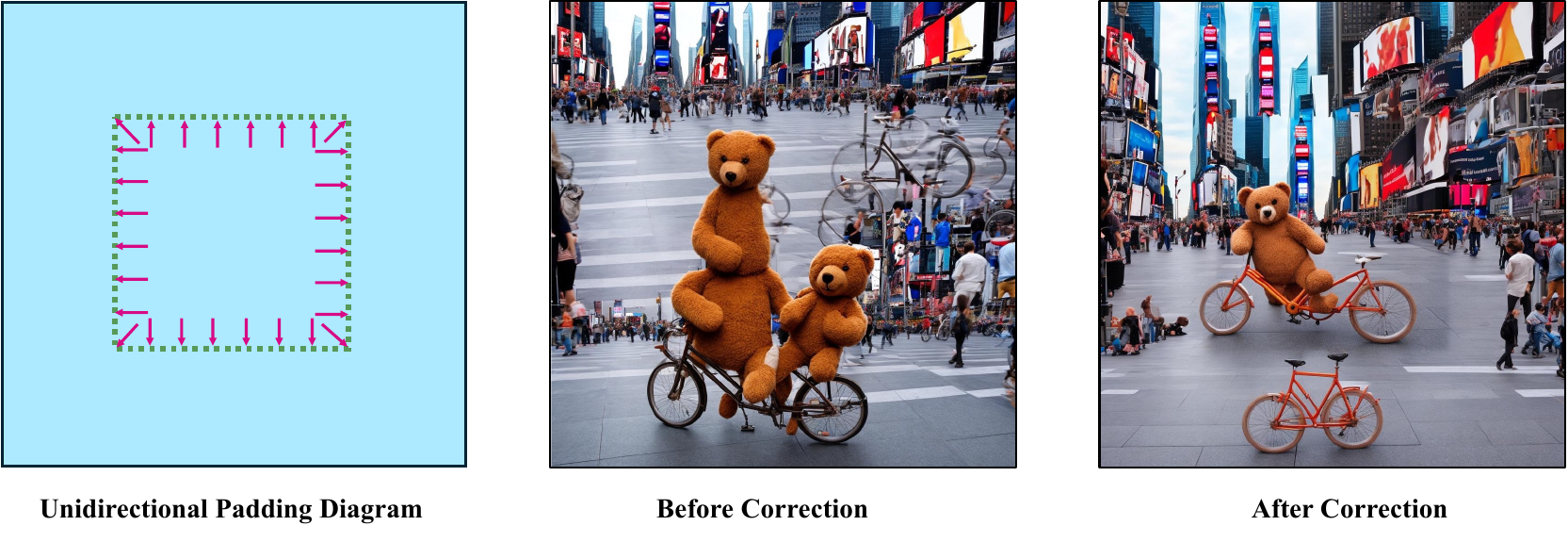}}
\vskip -0.1in
\caption{Position Information Correction. We applied unidirectional zero-padding to the central region of the feature map to facilitate faster propagation of position information. The images were generated at a resolution of 1024×1024, with the central region measuring 512×512. Prompt: \emph{A photo of a teddy bear riding a bike in Times Square.}}
\label{fig-analysis-3}
\end{center}
\vskip -0.3in
\end{figure}

\subsection{Position Encoding in Visual Neural Networks}
In transformer-based visual neural networks~\cite{dosovitskiy2020image, oquab2023dinov2}, explicit positional embeddings are incorporated into the attention mechanism, providing a clear source of positional information. In contrast, traditional convolution-based networks typically lack explicit positional embeddings. Although \citeauthor{islam2020much} demonstrated that convolutional networks inherently possess positional information closely related to the zero-padding mechanism, their work did not uncover how this positional information propagates or explicitly manifests within generative models, such as diffusion models.

\section{Analysis of the Position Encoding Mechanism}
In this section, we examine how the U-Net architecture establishes position encoding, specifically identifying which components and how they contribute to position information. Building on these analysis, we aim to deduce the cause of the disorder pattern in high-resolution image generation. All experiments in this section are conducted with \emph{SD-2.1.}

\subsection{Convolutional Layers Govern Position Encoding}
In traditional U-Net architectures, such as the U-Net in \emph{SD-1.X, SD-2.X, SD-XL}, and ~\emph{etc.}, no position-embedding is added to the latent tokens in attention operations, including both self-attention and cross-attention. Thus, the attention layers do not contain any position information. Aside from the attention layers, only convolution operations handle interactions among latent tokens, indicating that all position information comes from convolution layers. Inspired from~\citeauthor{islam2020much}, we probe that the position information originates from the zero-padding mechanism at the latent boundaries.

To validate this, we first design a trench-style zero-padding technique, which divides the latent feature into multiple convolution regions, as illustrated on the left side of Figure~\ref{fig-analysis-1}. Specifically, we apply bi-directional zero-padding to the trenches (splits) of the feature map during convolution operations. As shown in Figure~\ref{fig-analysis-1}, the added zero-padding effectively partitions the generated images. Regions with downward padding generate elements such as oceans and land, whereas regions with upward padding produce elements such as the sky. This demonstrates that zero-padding controls global positional consistency.

\begin{figure}[t]
\begin{center}
\centerline{\includegraphics[width=\columnwidth]{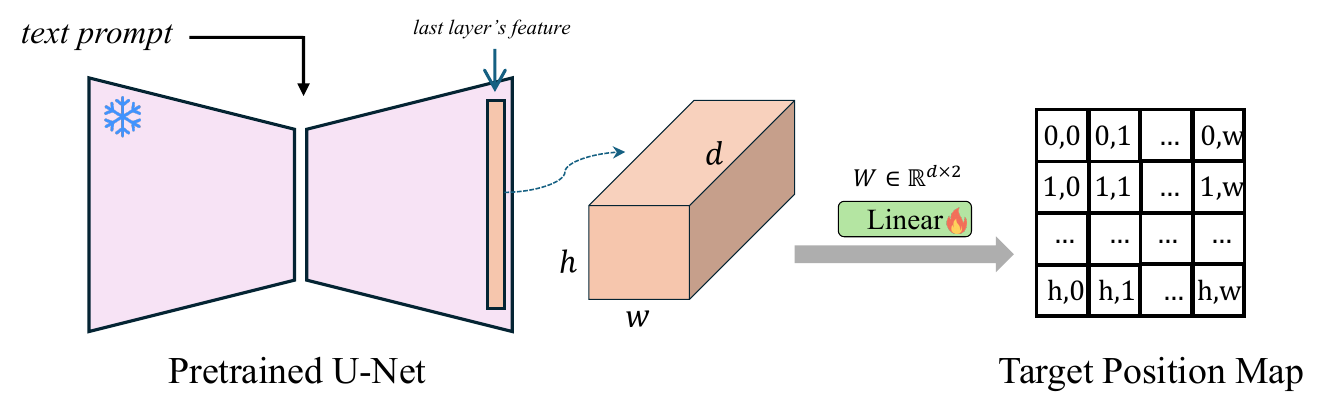}}
\vskip -0.1in
\caption{Position Information Quantification. We extract the feature from the last layer of U-Net and map it to a target position map with a trainable linear layer. The loss of the linear year reflects how much position information the feature contains.}
\label{fig-analysis-4}
\end{center}
\vskip -0.3in
\end{figure}

\begin{table}[t]
\caption{Loss values of the linear layer when training to map features to the target position map under different circumstances. Lower values indicate better position information encoding.}
\label{table-analysis}
\vskip -0.1in
\begin{center}
\begin{small}
\begin{tabular}{lcccr}
\toprule
Position Mode & Resolution & Loss\textdownarrow  \\
\midrule
Random Feature &  512$^2$ & 0.077 \\
Zero-padding & 512$^2$ & \textbf{0.006} \\
Circular-padding & 512$^2$ & 0.033 \\
\midrule
Zero-padding & 1024$^2$ & 0.023 \\
Circular-padding & 1024$^2$ & 0.053 \\
Dilated Convolution & 1024$^2$ & 0.023\\
PBC (ours) & 1024$^2$ & \textbf{0.017}\\
\midrule
Zero-padding & Central 512$^2$ in 1024$^2$ & 0.039 \\
Dilated Convolution & Central 512$^2$ in 1024$^2$ & 0.025\\
PBC (ours) & Central 512$^2$ in 1024$^2$ & \textbf{0.010}\\
\bottomrule
\end{tabular}
\end{small}
\end{center}
\vskip -0.1in
\end{table}

To further investigate the role of zero-padding, we replaced it with several common padding modes, including reflect padding, replicate padding, and circular padding. The results are shown in Figure~\ref{fig-analysis-2}, where only the zero-padded images exhibit a reasonable layout. This demonstrates that zero-padding not only maintains global positional consistency but also preserves image integrity.

\subsection{Insufficient Position Encoding in High-resolution Generation}
\label{analysis-2}
Building on the insight that zero-padding provides positional information, a key question arises: how do the tokens in the central regions obtain their positional information? We speculate that positional information propagates through stacked convolutional layers to cover the entire feature map. This hypothesis explains why high-resolution generation often results in disordered layouts - the propagation routes become longer at higher resolutions, causing insufficient positional encoding, particularly in the central regions.

To validate this hypothesis, we conduct an experiment aimed at improving positional information propagation in central regions by applying unidirectional zero-padding specifically at the center, as shown in Figure~\ref{fig-analysis-3}. This method successfully resolves the chaotic arrangement in the central region, correcting issues such as disorganized grounds and buildings, as well as the repetitive generation of bears.

\subsection{Quantitative Analysis for Position Information}
\label{analysis-3}
To confirm the above conclusions, we design an artful quantitative experiment aimed at exploring how much position information that U-Net can encode under different circumstances, as shown in Figure~\ref{fig-analysis-4}. We extract the last layer’s feature map from the U-Net at the first denoising time-step and map it to a target position map of the same size, containing only position coordinates. Only a simple trainable linear layer is used for this mapping, aimed at minimizing its fitting capability. After training the linear layer to convergence, the final loss reflects the extent of position information encoded in the feature. Greater position information results in a lower loss. The details and thorough ablation studies of the experiment strategies can be found in the Appendix.

The loss values under different circumstances are written in Table~\ref{table-analysis}. The top three rows compare random features, zero-padding features (original), and circular-padding features, confirming that U-Net encodes position information and zero-padding greatly attributes to it. The middle four rows present the loss values for high-resolution generation (1024$^2$). In the bottom three rows, we separate the central 512$^2$ region in the 1024$^2$ resolution to conduct the experiment for a fair comparison with the original resolution generation (top three rows). This confirms that the encoded positional information becomes fewer in the central region in high-resolution generation, validating our hypothesis in~\ref{analysis-2}. Dilated convolutions~\cite{he2023scalecrafter} alleviate this issue to some extent, but our method (described in the next section) achieves the best performance.

\section{Methods}
In this section, we introduce our training-free high-resolution generation method. 
\subsection{Preliminaries}
\myparagraph{Latent Diffusion Model (LDM)}
 A pre-trained LDM enables the transfer of standard Gaussian noise $z_T$ into an image latent $z_0$ aligned with a pre-given text prompt through reverse denoising processes. Here, the latent space $ \mathbf{Z} \subseteq \mathbb{R}^{h \times w \times c}  $ is characterized by a pre-trained Vector Quantized variational autoencoder (VQ-VAE). One can decode the image latent $z_0$ into real image $x$ by VAE decoder $x = \mathcal{D}(z_0)$.

The reverse denoising process gradually obtains less noisy latent $z_{t-1}$ from the noisy input $z_t$ at each timestep $t \in (1, T)$: 
\begin{equation}
p_\theta (z_{t-1}|z_t) = \mathcal{N}(z_{t-1};\mu_\theta(z_t, t),\Sigma_\theta(z_t, t))	
\end{equation}
where $\mu_\theta$ and $\Sigma_\theta$ are determined through a noise prediction network $\epsilon_\theta(z_t, t, C)$ with learnable parameter $\theta$. $C$ is the text prompt embedding. In traditional LDM architecture, the denoising network is realized using a time-conditional U-Net. 

\myparagraph{Problem Definition} Given a LDM with U-Net $\epsilon_\theta$ trained at the pre-defined fixed low-resolution $x\in \mathbb{R}^{H\times W\times 3}$, our objective is to synthesis high-resolution images $\tilde{x}\in \mathbb{R}^{\tilde{H}\times \tilde{W}\times 3}$ in a training-free manner. 
\subsection{Progressive Boundary Complement (PBC)}
One straightforward approach to enhance the propagation of position information is to insert additional real boundaries inside the feature map, as we discussed in \ref{analysis-2}. In that experiment,  we added a unidirectional zero-padding square at the central region of the feature map. However, though it solves the disorder in the central region, this approach will cause the visible split shown at the square edges, we define it as the split effect, as shown in the ``After Correction" part of Figure~\ref{fig-analysis-3}.

\begin{figure}[t]
\begin{center}
\centerline{\includegraphics[width=\columnwidth]{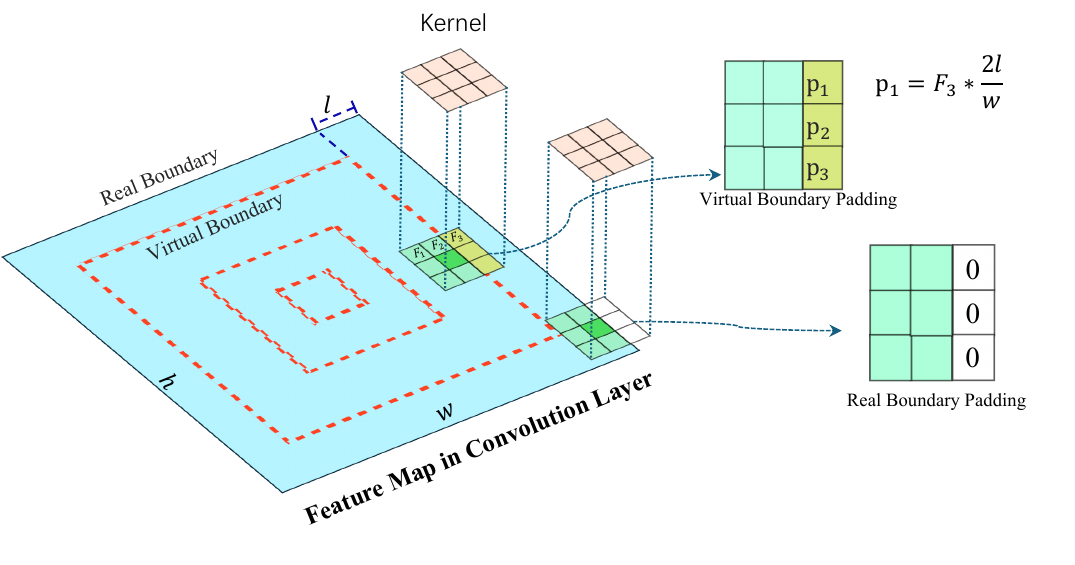}}
\vskip -0.1in
\caption{Progressive Boundary Complement (PBC) for training-free high-resolution image generation. PBC enhances the propagation of position information by inserting a series of designed virtual boundaries into the feature map.}
\label{fig-method-1}
\end{center}
\vskip -0.1in
\end{figure}
\subsubsection{Virtual Boundary}
To address the issue of discontinuity, we propose the Virtual Boundary technique, which aims to construct internal boundaries within the feature maps. This approach effectively mitigates the discontinuities introduced by zero-padding while preserving the propagation of positional information.

The cause of the split effect can be easily inferred. Tokens located just inside the zero-padding square cannot gather information from the tokens right outside it during convolution operations, leading to incorrect feature aggregation at the edges of the square. Thus, we proposed a novel padding trick, named valued-padding, designed to replace zero-padding with a value that contains features from the outside token to enhance the feature aggregation. On the other hand, we expect it to have the zero-padding's property to provide the position information. Therefore, we set the padding value as a proportional of the outside token's value, formed as:
\begin{equation}
	p = F * \lambda, \quad \lambda \in [0,1]
\end{equation}
where  $p$ is the padding value,  $F$  is the feature (value) of the outside token, and  $\lambda$  is the ratio. Ideally, a lower  $\lambda$  results in the virtual boundary providing stronger position information, as it more closely resembles zero-padding, we prove it in the Sec. \ref{ablation-1}.

\myparagraph{Random Boundary Perturbation} While valued-padding helps alleviate the split effect, it does not completely eliminate discontinuities at boundary edges. To further smooth transitions, we introduce random perturbations to the positions of the virtual boundaries within a small range across different convolutional layers:
\begin{equation}
	\tilde{l} = l + \delta, \quad \delta \sim \text{U}(-r, r)
\end{equation}
where $l$ is the distance between boundary center to the real image boundary. $\tilde{l}$ is the perturbed distance, $r$ denotes the perturbation range.

\subsubsection{Hierarchical Virtual Boundaries Placement}
As discussed in Section \ref{analysis-3}, position information originates from the feature boundaries and propagates through stacked convolution layers, which results in stronger position encoding near the edges while gradually diminishing towards the center.  Therefore, to simulate this trend, we discretely place multiple virtual boundaries within the feature map, ensuring that those closer to the feature boundaries have a stronger ability to provide position information. 

Specifically, we first define the padding ratio as  $\lambda = \frac{2l}{s}$, where  $l$  represents the distance from the virtual boundary to the real boundary of the feature map, and  $s$  denotes the width or height of the feature map, depending on the orientation of the virtual boundary. This formulation ensures that virtual boundaries closer to the feature edge have larger ratios, effectively capturing stronger positional information. For simplicity, we refer to the virtual boundary as $V_\lambda$, indexed by its ratio $\lambda$.

\begin{table*}[t]
\caption{Image quantitative comparisons with baselines. We evaluate 2048×2048 resolution images generated by \emph{SD-XL}. Our method achieves the best or second-best scores for all quality-related metrics among all training-free baselines with lower additional time costs. The best result are marked in \textbf{bold}, and the second best results are marked by \underline{underline}. $*$ indicates the method needs further fine-tuning of \emph{SD-XL}.} 
\label{table-exp-1}
\vskip -0.1in
\begin{center}
\begin{small}
\begin{tabular}{lcccccc}
\toprule
Method & KID \textdownarrow &IS \textuparrow & HPS \textuparrow & ImageReward \textuparrow& Content Richness \textdownarrow & Inference Time \textdownarrow \\
\midrule
DI & 0.0092 & 9.25& 20.76&0.645&30.30&68s\\
ScaleCrafter~\cite{he2023scalecrafter} &\underline{0.0067}&	\underline{9.53}&	\underline{21.15}&	\textbf{1.078}&	29.50& \underline{71s} (+4\%)\\
FouriScale~\cite{huang2025fouriscale} &0.0103&9.38	&21.13	&1.023&	\underline{29.61}& 130s (+91\%)\\
PBC (ours) &\textbf{0.0067}& \textbf{9.95}& 	\textbf{21.20}&  \underline{1.036}& 	\textbf{29.42}& \textbf{71s} (+4\%)\ \\
\midrule
DemoFusion$*$~\cite{du2024demofusion} &0.0088	&11.07&	21.33	&1.090&	29.47&189s (+178\%)\\
\bottomrule
\end{tabular}
\end{small}
\end{center}
\vskip -0.1in
\end{table*}

Secondly, we define a $N$ discrete hierarchical virtual boundaries extending from the real boundary toward the image center, denoted as: 
\begin{equation}
\mathcal{V}_N = \{V_{\lambda_n} \mid \lambda_n = \frac{n}{N+1},  n = 1, 2, \dots, N\}
\end{equation}
where  $\lambda_n$  represents the ratio for each virtual boundary. 

The overall process of our Progressive Boundary Complement (PBC) is shown in Figure~\ref{fig-method-1}.
\begin{figure}[t]
\begin{center}
\centerline{\includegraphics[width=\columnwidth]{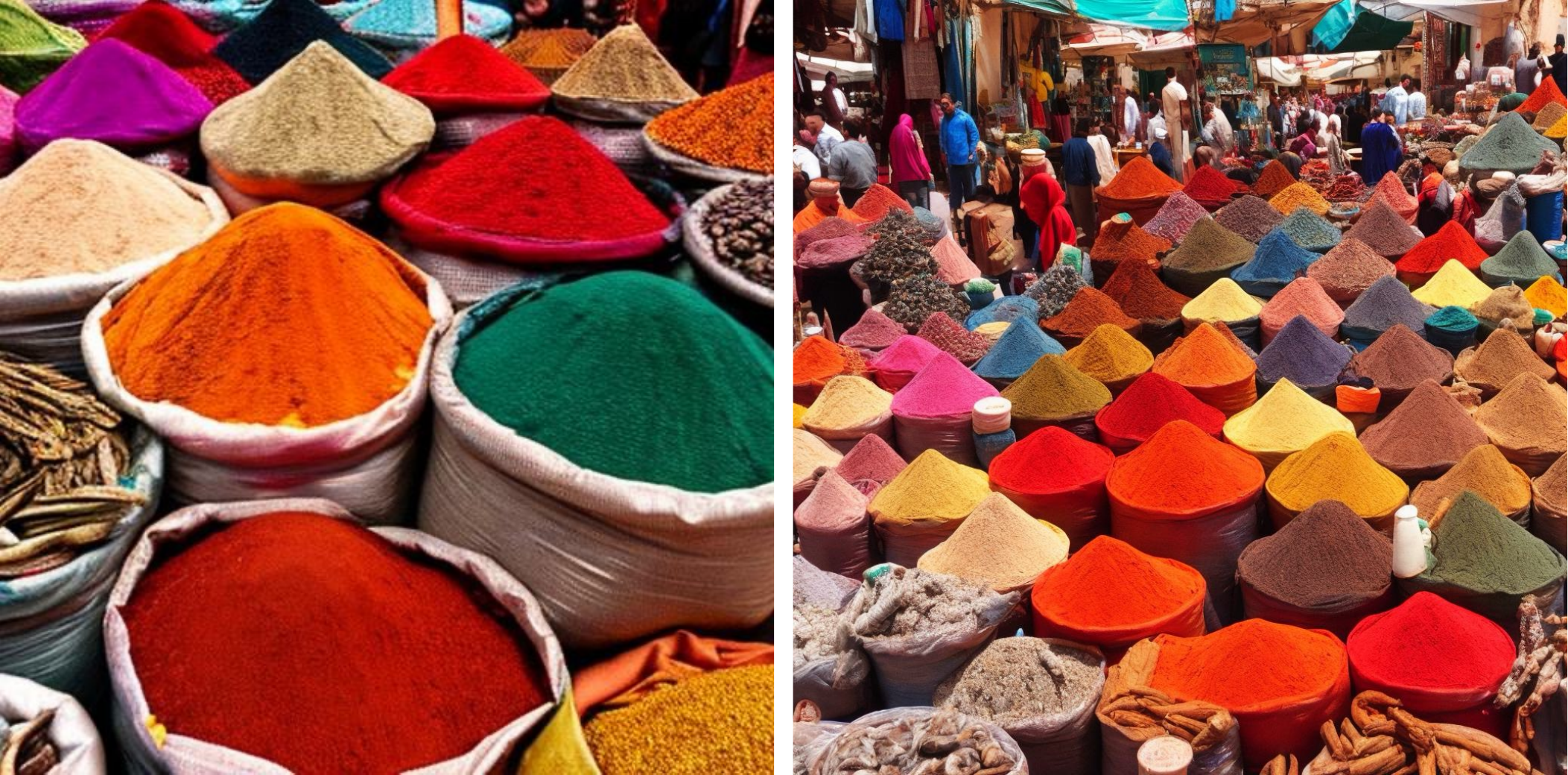}}
\vskip -0.1in
\caption{Content-enriched Image Generation. \emph{left}: original 512×512 generation, \emph{right}: 1024×1024 generation using PBC. Our method enhances content richness, introducing additional details such as more grain stacks and a distant crowd of people.  Prompt: \emph{A lively outdoor market in Morocco, full of colorful spices and fabrics.} Model: \emph{SD-2.1}.}
\label{fig-method-2}
\end{center}
\vskip -0.3in
\end{figure}

\begin{figure*}[t]
\begin{center}
\centerline{\includegraphics[width=\textwidth]{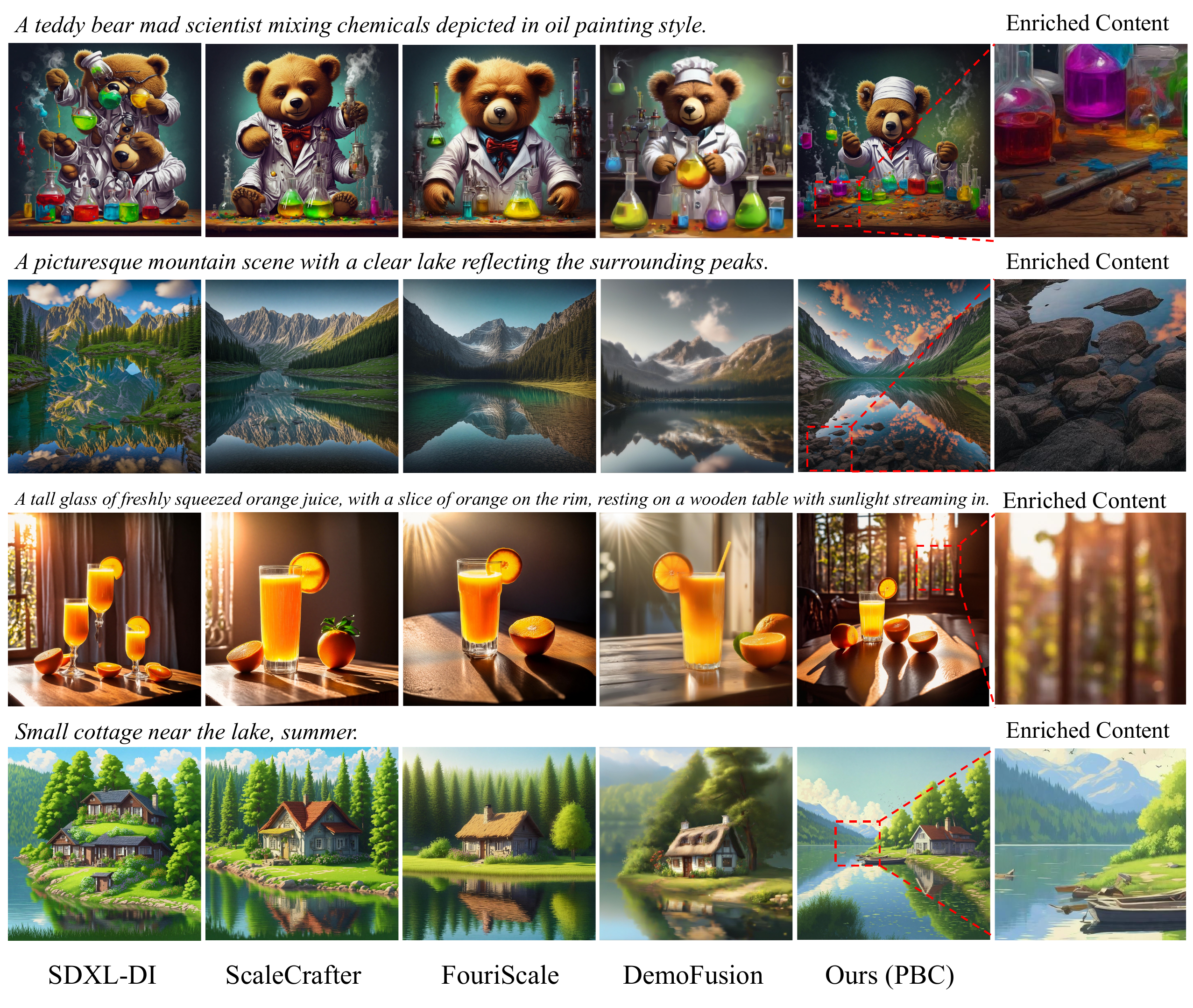}}
\vskip -0.1in
\caption{PBC is compared to baseline methods at a resolution of 2048×2048 using \emph{SD-XL}. Our approach demonstrates the ability to generate images with enriched content and high-quality results. Zoom-in for best view.}
\label{fig-exp-1}
\end{center}
\vskip -0.3in
\end{figure*}

\subsection{Image Generation with Enriched Content}
Most high-resolution generation methods primarily aim to produce images that closely resemble those at the training resolution but on a larger scale, aligning to some extent with the objective of super-resolution. In other words, these methods essentially generate an upscaled version of the training-resolution image without introducing new semantic content. In contrast, our proposed Progressive Boundary Complement (PBC) method leverages virtual boundaries to effectively expand the image canvas, facilitating the synthesis of high-resolution images with enriched and additional content, as illustrated in Figure~\ref{fig-method-2}.

\myparagraph{Content Richness Metric} 
Since existing image generation evaluation metrics do not accurately quantify the richness of content in generated images, we introduce a novel metric, Content Richness (CR), to assess the diversity and richness of details in generated images. Specifically, given a generated image  $x \in \mathbb{R}^{H \times W \times 3}$ , we first partition it into  $k^2$ ($k=3$ in our experiment) equally sized image patches, denoted as  $\{x_1, x_2, \dots, x_{k^2}\}$ , where each patch satisfies  $x_i \in \mathbb{R}^{(H/k) \times (W/k) \times 3}$ . Next, we compute the pairwise CLIP ~\cite{radford2021learning} similarity between all image patches. Let  $f_i$ denote the CLIP feature of patch $x_i$. The overall similarity score is obtained by computing the sum of pairwise cosine similarity across all patch pairs
\begin{equation}
S = \sum_{i=1}^{k^2} \sum_{\substack{j=1 \ j \neq i}}^{k^2} \frac{f_i \cdot f_j}{| f_i | | f_j |}.
\end{equation}
A lower similarity score  $S$  indicates higher content diversity within the image.

\section{Experiments}
\subsection{Experimental setup}
\myparagraph{Implementation Details}
We conduct experiments based on the open-source latent diffusion model \emph{SD-XL}. The hyper-parameter $N$, representing the number of virtual boundaries, is set to 31. We adhere to the default settings and utilize a 50-step DDIM~\cite{song2020denoising} sampling process. Following~\cite{he2023scalecrafter}, our PBC strategy is applied only during the first 25 denoising steps, with FreeU~\cite{si2024freeu} integrated. We implement our method by unfold-fold technique of convolution, see details in Appendix. All experiments are conducted using a single RTX 3090 GPU.

\myparagraph{Dataset and Evaluation Metrics}
For test prompt, we randomly select 500 high-quality prompts from Laion-5B~\cite{schuhmann2022laion}, each prompt is evaluated with 10 seeds.
To evaluate the quality of the generated images, we report the following metrics: Kernel Image Distance (KID)~\cite{binkowski2018demystifying}, which measures the semantic similarity between the generated high-resolution images and the original resolution images; Inception Score (IS)~\cite{salimans2016improved}, which assesses the diversity of the generated images; and Human Preference Score (HPS)~\cite{wu2023better} along with ImageReward~\cite{xu2024imagereward}, both of which gauge human preferences for the generated images.
\begin{figure*}[t]
\begin{center}
\centerline{\includegraphics[width=\textwidth]{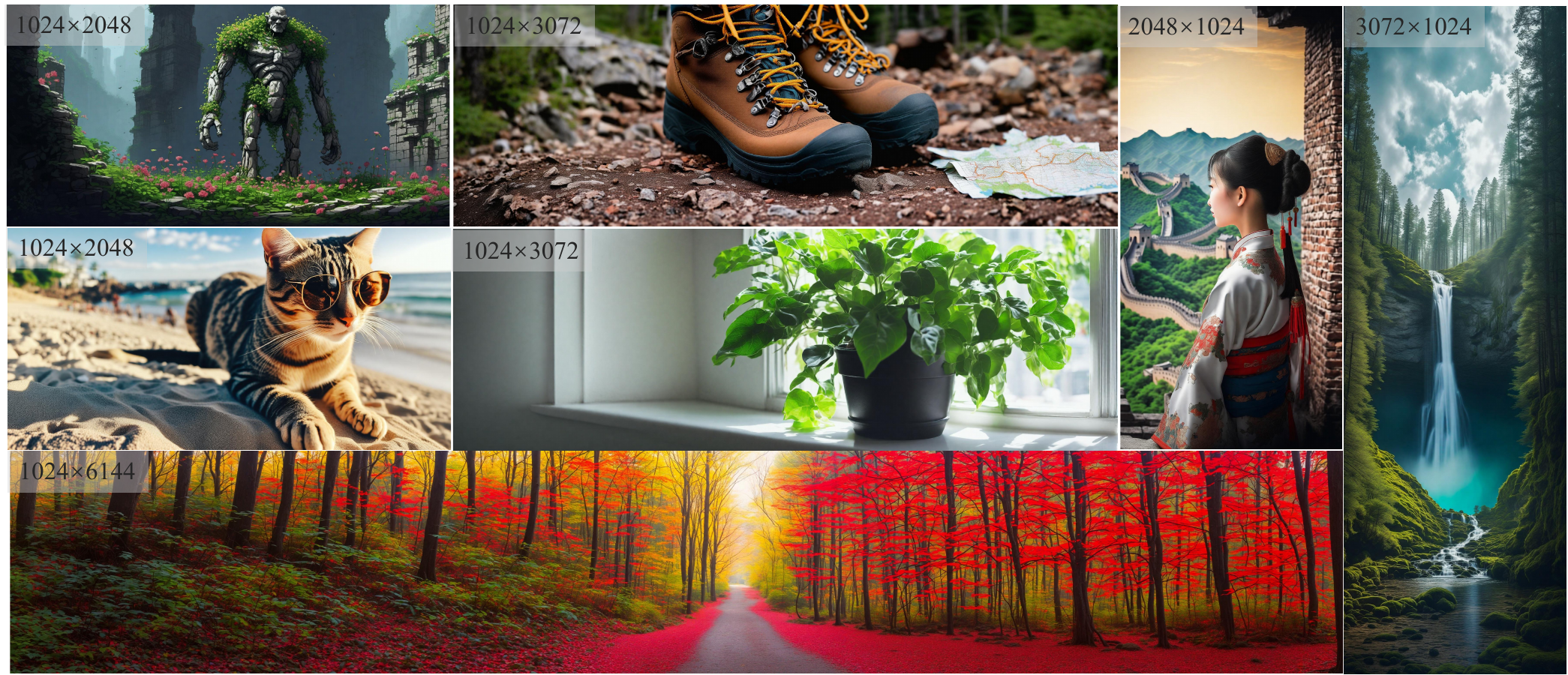}}
\vskip -0.1in
\caption{PBC is able to generate non-square images. The images are generated by \emph{SD-XL}. Zoom-in for best view.}
\label{fig-exp-3}
\end{center}
\vskip -0.2in
\end{figure*}

\myparagraph{Baselines} 
We compare our methods with other high-resolution image generation methods. Training-free: (i) \emph{SD-XL}~\cite{rombach2022high} direct inference (DI) (ii) \emph{ScaleCrafter}~\cite{he2023scalecrafter} (iii)  \emph{FouriScale}~\cite{huang2025fouriscale} Training-based: (iv) \emph{DemoFusion}~\cite{du2024demofusion}. 

\subsection{Evaluation}
\myparagraph{Quantitative Results}
The quantitative results confirm the superiority of our method, which are shown in Table~\ref{table-exp-1}. Our method achieves the best or second-best scores for all quality-related metrics among all training-free baselines with lower additional time costs. Our method achieves best Content Richness (CR) results among all the baselines, which demonstrate our method can produce content-enriched images.

\begin{figure*}[t]
\begin{center}
\centerline{\includegraphics[width=\textwidth]{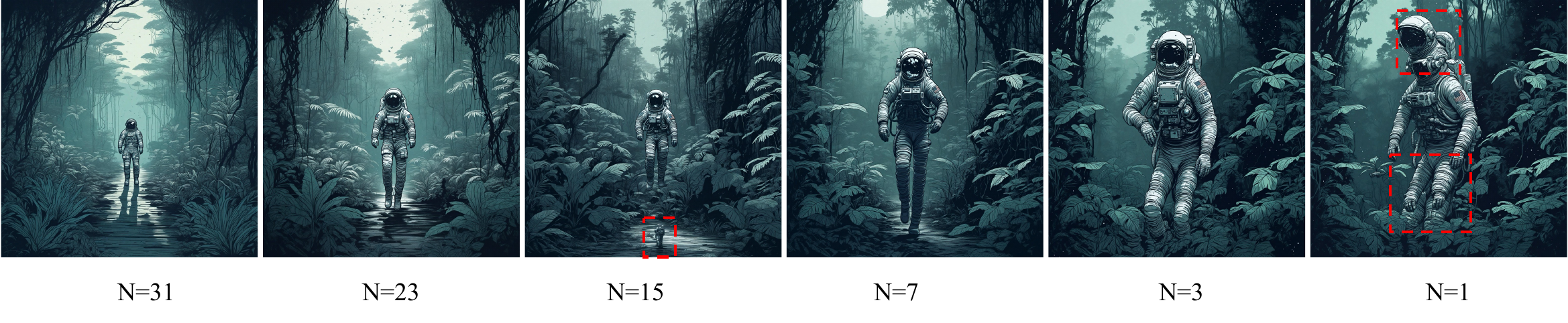}}
\vskip -0.1in
\caption{Effect of virtual boundaries numbers on image quality and content richness. A greater number of virtual boundaries leads to higher image quality and richer content diversity. The images are generated by \emph{SD-XL} with resolution 2048$^2$.}
\label{fig-exp-2}
\end{center}
\vskip -0.3in
\end{figure*}

\begin{figure}[ht]
\begin{center}
\centerline{\includegraphics[width=\columnwidth]{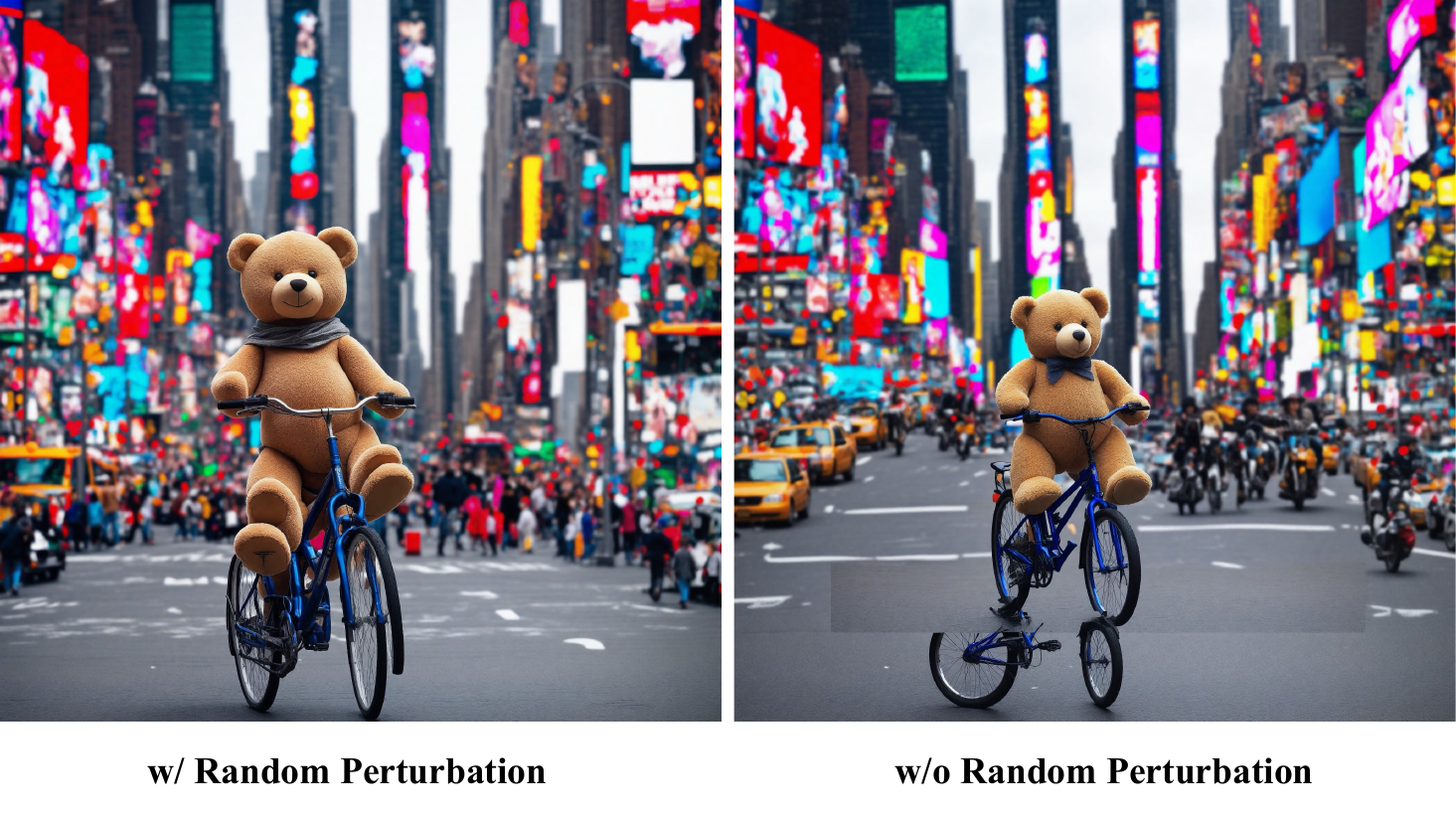}}
\caption{Effect of Random Perturbation of Virtual Boundary with $N = 3$. Random perturbation mitigate the split effect.}
\label{fig-exp-4}
\end{center}
\vskip -0.3in
\end{figure}
\begin{table}[t]
\caption{The virtual boundary ratio $\lambda$ to the position information.}
\label{table-ablation-1}
\vskip -0.1in
\begin{center}
\begin{small}
\begin{tabular}{@{}ccccccc@{}}
\toprule
$\lambda$ &0 & 0.2 & 0.4 & 0.6 & 0.8 & 1 \\
\midrule
Loss & 0.0200 & 0.0187 & 0.0176 & 0.0172 & 0.0189 & 0.0230\\
\bottomrule
\end{tabular}
\end{small}
\end{center}
\vskip -0.3in
\end{table}

\myparagraph{Qualitative Results}
The qualitative results between our method and baselines are shown in Figure~\ref{fig-exp-1}. The direct inference of \emph{SD-XL} often cause repetitive local image elements and disordered patterns. ScaleCrafter also tends to produce localized repetitions (see the first line in the figure). FouriScale delivers high-quality visual results, but the image content is relatively limited. The images generated by DemoFusion exhibit noticeable blurriness. Our method not only generates high-quality images but also produces content-rich scenes. For instance, in the fourth row, our method generates additional boats and mountains.

\myparagraph{Additional Qualitative Results} Our method can generate non-square high-resolution images by adding virtual boundaries in only one direction, the generated results are shown in the Figure~\ref{fig-exp-3}.

\begin{table}[t]
\caption{Effect of the virtual boundary numbers $N$ on position information, represented by the loss values of the mapping linear layer.}
\label{table-ablation-2}
\vskip -0.1in
\begin{center}
\begin{small}
\begin{tabular}{lccccc}
\toprule
$N$ &0 & 1 & 3 & 5 & 7 \\
\midrule
Loss& 0.0230 & 0.0173 & 0.0177 & 0.0176 & 0.0175\\
\bottomrule
\end{tabular}
\end{small}
\end{center}
\vskip -0.2in
\end{table}

\subsection{Ablation Study}
\myparagraph{Effect of Virtual Boundary and the Random Perturbation}
\label{ablation-1}
We first conduct ablation studies on the virtual boundary ratio $\lambda$ , following a similar experimental setup as in Section~\ref{analysis-3}. Specifically, we place a single virtual boundary within the feature map and evaluate the overall positional information retained under different  $\lambda$  values. The results, presented in Table~\ref{table-ablation-1}, indicate that the highest overall positional information is achieved when  $\lambda = 0.6$. This is because smaller  $\lambda$  values restrict the propagation of external position information, whereas larger  $\lambda$  values reduce the inherent position information provided by the virtual boundary itself.

Next, we evaluate the effectiveness of the random perturbation technique, as shown in Figure~\ref{fig-exp-4}. For better visualization, we set the number of virtual boundaries to  $N=3$ . Without random perturbation, a clear split effect is observed.

\myparagraph{Ablation Study on Virtual Boundary Numbers}
To assess the impact of the number of virtual boundaries  $N$ , we visualize the generated images at a 2048² resolution using SD-XL under different values of  $N$, as shown in Figure~\ref{fig-exp-2}. As  $N$ increases, the images exhibit richer content and larger scene compositions, indicating that additional virtual boundaries effectively expand the perceived image boundaries. Conversely, an insufficient number of virtual boundaries leads to inconsistencies in image details, resulting in fragmented or incoherent visual elements.

Similarly, we also evaluate the impact of  $N$  on positional information within the generated images, as shown in Table~\ref{table-ablation-2}. The total amount of positional information does not change significantly as $N$  increases. This is because, while additional virtual boundaries contribute new position information, they also partially obstruct the propagation of external position information.
\section{Conclusion}
In this paper, we explore the position encoding mechanism in diffusion U-Net and identify inconsistent positional encoding as the root cause of repetitive and disordered patterns in high-resolution image generation. To address this, we propose Progressive Boundary Complement (PBC), a training-free method that enhances positional information propagation via hierarchical virtual boundaries. PBC effectively mitigates spatial inconsistencies, expands image boundaries, and enriches content diversity. Extensive experiments demonstrate its superiority over existing methods in both image quality and content richness, offering a novel perspective for improving high-resolution generation in diffusion models.
\bibliography{main}
\bibliographystyle{icml2025}
%%%%%%%%%%%%%%%%%%%%%%%%%%%%%%%%%%%%%%%%%%%%%%%%%%%%%%%%%%%%%%%%%%%%%%%%%%%%%%%
%%%%%%%%%%%%%%%%%%%%%%%%%%%%%%%%%%%%%%%%%%%%%%%%%%%%%%%%%%%%%%%%%%%%%%%%%%%%%%%
% APPENDIX
%%%%%%%%%%%%%%%%%%%%%%%%%%%%%%%%%%%%%%%%%%%%%%%%%%%%%%%%%%%%%%%%%%%%%%%%%%%%%%%
%%%%%%%%%%%%%%%%%%%%%%%%%%%%%%%%%%%%%%%%%%%%%%%%%%%%%%%%%%%%%%%%%%%%%%%%%%%%%%%
\newpage
\appendix
\onecolumn
\section{Details and Ablation Studies on Position Information Quantification}
\subsection{Experiment Details}
All position information quantification in the main manuscript follows the same experimental setup for a fair comparison. We extract the feature maps from the last convolutional layer of U-Net in \emph{SD-2.1} in the first denoising step, using the prompt ``An Asian girl smiling.” with a fixed random seed. The linear layer is trained for 50,000 iterations using the Adam optimizer with an initial learning rate of 0.0001. The values in the target position map are linearly normalized to $[0,1]$, and Mean Squared Error (MSE) loss is used as the objective loss function.
\subsection{Ablation Studies}
We conducted the following ablation study to validate the rationality of our experimental setup, comparing the results of zero-padding and circular-padding with original resolution under different configurations.
\subsubsection{Ablations Study on the Denoising Step}
We first conduct an ablation study to determine the optimal denoising step for position information quantification, as shown in the table below. The loss difference between zero-padding and circular-padding gradually shrinks to zero by the 49th step. This aligns with findings from ~\cite{yi2024towards}, which suggest that structural shaping or element positioning primarily occur during the early stages of the diffusion denoising process. This indicates that zero-padding contributes minimally to position encoding in the final denoising steps. Based on this observation, we select the initial denoising stage as the basis for our quantitative experiments.
\begin{table}[h]
\caption{Ablations Study on the Denoising Step.}
\begin{center}
\begin{small}
\begin{tabular}{cccc}
\toprule
& 0-step &25-step& 49-step \\
\midrule
Zero-padding& 0.0063 &0.0153&0.0197\\
Circular-padding& 0.0333 &0.0295&0.0197\\
\bottomrule
\end{tabular}
\end{small}
\end{center}
\end{table}
\subsubsection{Ablations Study on the U-Net blocks}
Second, we conduct an ablation study to determine the optimal U-Net block layer for position information quantification, as shown in the table below. The first block contributes minimally to positional encoding, as zero-padding information only propagates through two convolution layers. On the other hand, the middle block has a significantly reduced feature map resolution, allowing the linear layer to fully fit the position mapping, making it less informative for evaluating position encoding effects.
\begin{table}[h]
\caption{Ablations Study on the U-Net blocks.}
\begin{center}
\begin{small}
\begin{tabular}{cccc}
\toprule
& first-block &middle-block& last-block\\
\midrule
Zero-padding& 0.0507 &0.000&0.0063\\
Circular-padding& 0.0637 &0.000&0.0333\\
\bottomrule
\end{tabular}
\end{small}
\end{center}
\end{table}
\subsubsection{Ablations Study on the Prompts}
We select three different prompts for comparison: “An Asian girl smiling.” (girl), “A beautiful London city.” (city), and “A lovely dog running.” (dog). The results, presented in the table below, show no significant differences across the different prompts.

\begin{table}[h]
\caption{Ablations Study on the Prompts.}
\begin{center}
\begin{small}
\begin{tabular}{cccc}
\toprule
& dog &city& girl\\
\midrule
Zero-padding& 0.0100 &0.089&0.0063\\
Circular-padding& 0.0345 &0.0476&0.0333\\
\bottomrule
\end{tabular}
\end{small}
\end{center}
\end{table}
\subsection{Ablation Study on Padding Type}
Finally, we conduct ablation study on the effect of padding type to position information quantification, as shown in the table below.
\begin{table}[h]
\caption{Ablations Study on the Padding-type.}
\begin{center}
\begin{small}
\begin{tabular}{ccccc}
\toprule
& zero-padding &reflect-padding& replicate-padding & circular-padding\\
\midrule
Loss \textdownarrow & \textbf{0.0063}& 0.0357 & 0.0222 & 0.0333\\
\bottomrule
\end{tabular}
\end{small}
\end{center}
\end{table}
\section{Implementation of Progressive Boundary Complement (PBC)}
As we mentioned in the main manuscript, we leverage the unfold-fold to implement our method, the pseudo-code is as follows.
\begin{algorithm}[h]
\caption{Progressive Boundary Complement (PBC) via Unfold-Fold Convolution}
\label{alg:PBC}
\begin{algorithmic}[1]
\State \textbf{Input:} Feature map $\mathbf{F} \in \mathbb{R}^{B \times C \times H \times W}$
\State \textbf{Input:} Precomputed virtual boundary indices $\mathcal{V} = \{V_1, V_2, \dots, V_L\}$
\State \textbf{Input:} Corresponding ratio multipliers $\Lambda = \{\lambda_1, \lambda_2, \dots, \lambda_L\}$
\State \textbf{Output:} Modified feature map $\tilde{\mathbf{F}}$
\State
\State \textbf{Step 1: Unfold Feature Map}
\State $\mathbf{U} \gets \text{Unfold}(\mathbf{F}, K, S)$ 
\Comment{Extract patches with kernel size $K$ and stride $S$}
\State
\State \textbf{Step 2: Apply Virtual Boundary Scaling}
\For{$\ell = 1$ \textbf{to} $L$} 
    \State $\mathbf{U}[:, :, V_\ell] \gets \mathbf{U}[:, :, V_\ell] \times \lambda_\ell$ 
    \Comment{Scale patches corresponding to virtual boundaries}
\EndFor
\State
\State \textbf{Step 3: Fold Back to Original Structure}
\State $\tilde{\mathbf{F}} \gets \text{Fold}(\mathbf{U}, H, W, K, S)$ 
\Comment{Reconstruct the feature map from patches}
\State
\State \textbf{Return} $\tilde{\mathbf{F}}$
\end{algorithmic}
\end{algorithm}
\section{Additional Qualitative Results}
We present additional Qualitative Results below.
\begin{figure*}[t]
\begin{center}
\centerline{\includegraphics[width=\textwidth]{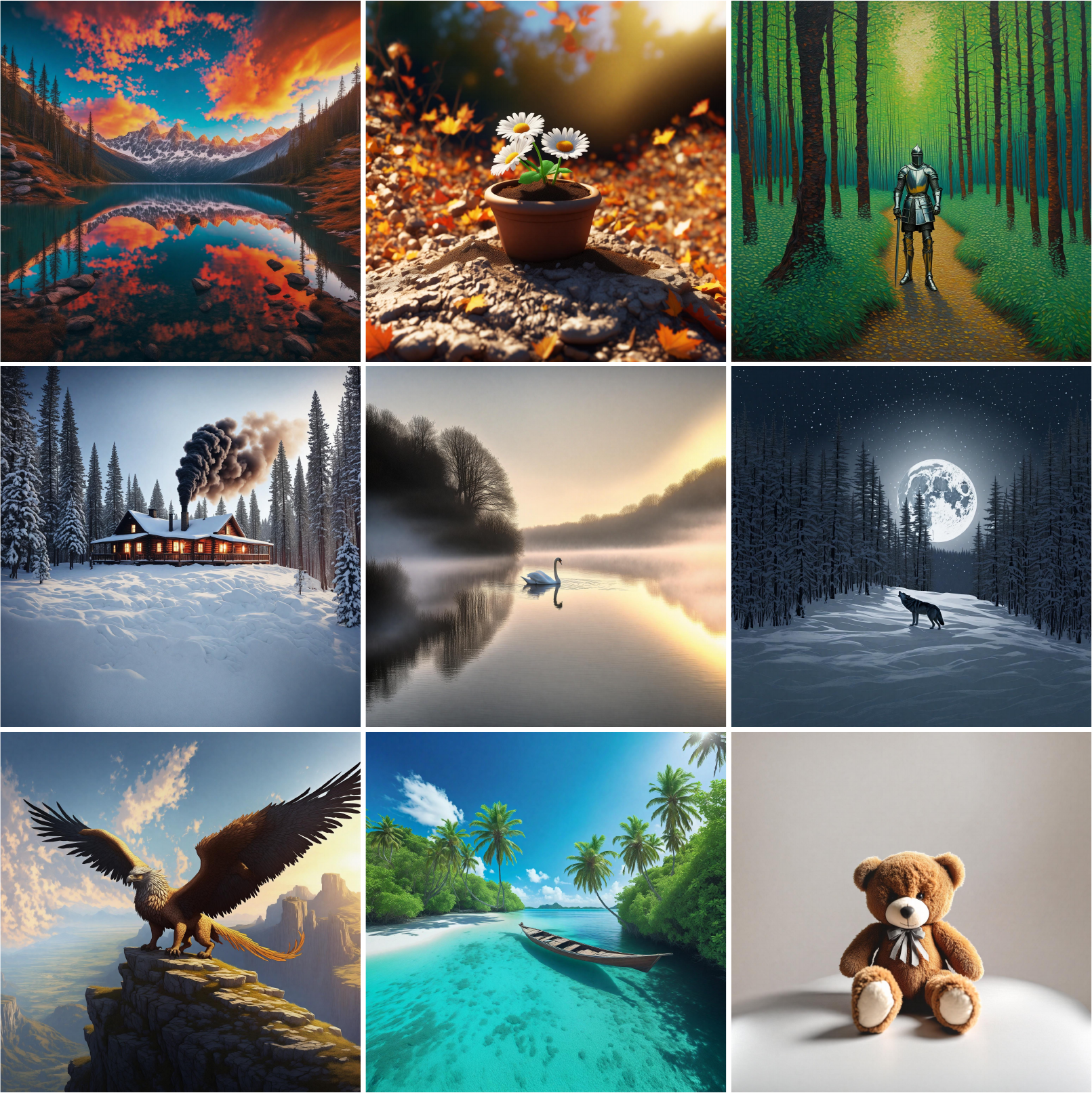}}
\vskip -0.1in
\caption{Square images with 2048$^2$ resolution by \emph{SD-XL}. Zoom in for best view.}
\end{center}
\vskip -0.2in
\end{figure*}

\begin{figure*}[t]
\begin{center}
\centerline{\includegraphics[width=\textwidth]{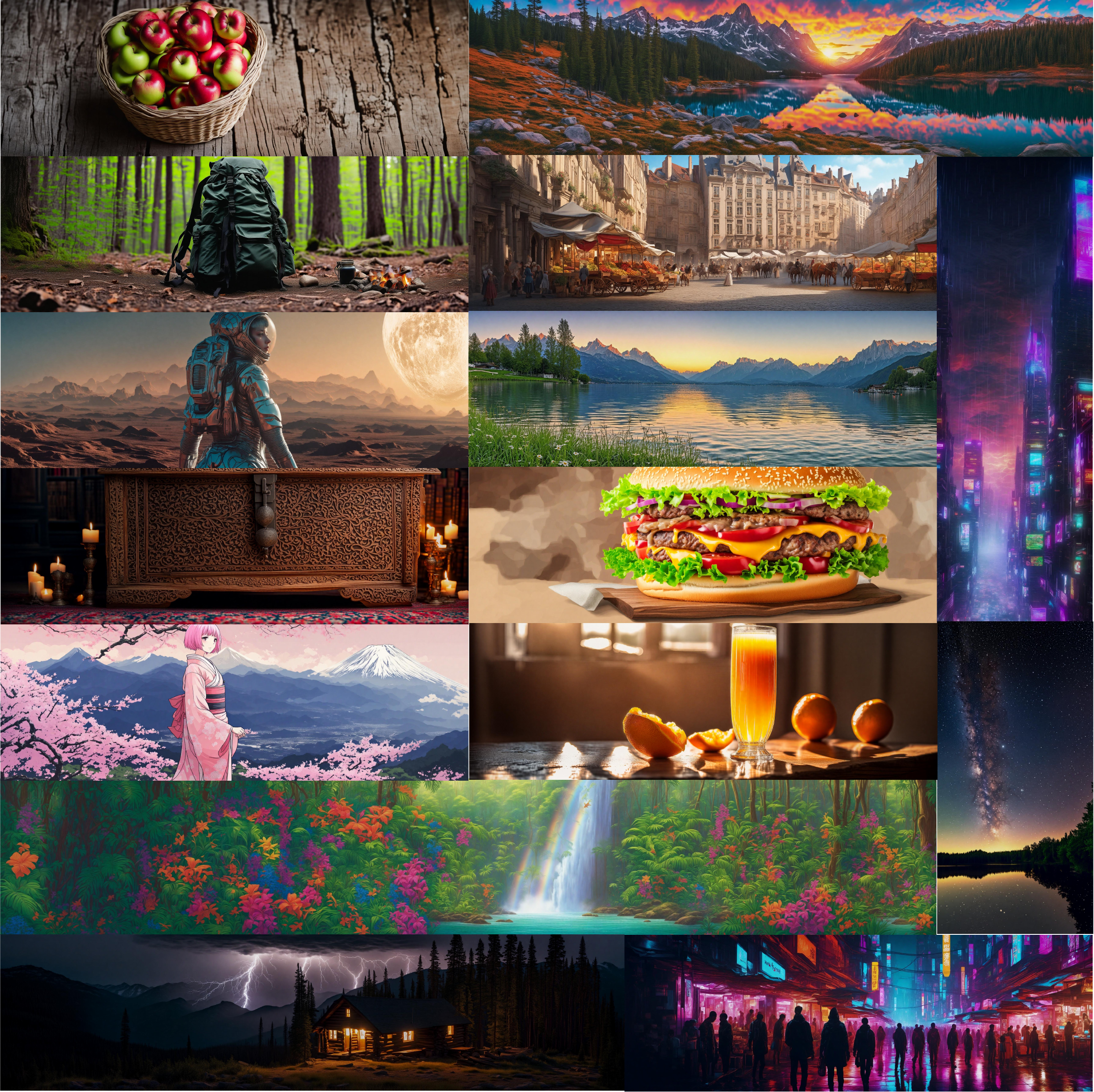}}
\vskip -0.1in
\caption{Non-square images by \emph{SD-XL}. Zoom in for best view.}
\end{center}
\vskip -0.2in
\end{figure*}
\end{document}